\documentclass[conference]{IEEEtran}
\IEEEoverridecommandlockouts
\usepackage{cite}
\usepackage{amsmath,amssymb,amsfonts}
\usepackage{algorithm}
\usepackage{algpseudocode}
\usepackage{bbm}
\usepackage{graphicx}
\usepackage{textcomp}
\usepackage{xcolor}
\usepackage{tabularx}
\usepackage{array}
\usepackage{subcaption}
\def\BibTeX{{\rm B\kern-.05em{\sc i\kern-.025em b}\kern-.08em
    T\kern-.1667em\lower.7ex\hbox{E}\kern-.125emX}}

\begin{document}

\title{Bayesian-Optimized Superpixel-GrabCut for Traceable Optic Disc Segmentation\\
}

\author{\IEEEauthorblockN{1\textsuperscript{st} Shraddha Changune}
\IEEEauthorblockA{\textit{Electrical Engineering} \\
\textit{Indian Institute of Technology}\\
Delhi, India \\
eet242394@iitd.ac.in}
\\
\IEEEauthorblockN{3\textsuperscript{rd} Gautam Das}
\IEEEauthorblockA{\textit{Electronics and Communication Engineering} \\
\textit{Gauhati University}\\
Guwahati, India \\
gautamdas2004@gmail.com}
\and
\IEEEauthorblockN{2\textsuperscript{nd} Vivek Noel Soren}
\IEEEauthorblockA{\textit{Electrical Engineering} \\
\textit{Indian Institute of Technology}\\
Delhi, India \\
eez248408@iitd.ac.in}
\\
\IEEEauthorblockN{4\textsuperscript{th} Tapan Kumar Gandhi}
\IEEEauthorblockA{\textit{Electrical Engineering} \\
\textit{Indian Institute of Technology}\\
Delhi, India \\
tgandhi@iitd.ac.in}
}


\maketitle

\begin{abstract}
Optic disc (OD) segmentation is essential for diagnosing ophthalmic pathologies from retinal fundus images. However, prevailing deep learning approaches operate as opaque black boxes, lacking the inference-stage mathematical traceability---a critical requirement for algorithmic auditing and failure analysis in clinical workflows. This paper presents a fully algorithmically traceable and trainable segmentation pipeline that jointly combines superpixel decomposition, hybrid brightness-proximity superpixel scoring, morphological regularization, iterative GrabCut refinement, and elliptical shape fitting. The hyperparameter optimization is formulated as an objective function and solved via Bayesian optimization to eliminate manual parameter tuning. A quantitative evaluation on the Drishti-GS dataset demonstrates that our method achieves a Dice coefficient of 0.9536, matching state-of-the-art performance. By maintaining explicit mathematical transparency across all processing stages, our framework offers a deterministic, traceable alternative to black-box architectures for medical review and debugging.
\end{abstract}


\section{Introduction}
The optic disc (OD) is a foundational anatomical landmark within the ocular fundus. Appearing in retinal fundus photographs as a bright, elliptical region approximately 1.5 mm in diameter, it marks where retinal ganglion cell axons converge to form the optic nerve\cite{chan2018role}. Accurate tracking of the OD boundary is essential for diagnosing ophthalmic and systemic conditions, particularly glaucoma-the second leading cause of irreversible blindness worldwide\cite{Kimetal2022}.\\
Traditional OD segmentation methods fall into non-model-based (thresholding \cite{aquino2010detecting}, clustering \cite{cheng2013superpixel}) and model-based (template matching \cite{lalonde2002fast}, deformable contours\cite{haleem2018novel}) paradigms. The former are fast but fragile under vessel occlusion; the latter incorporate shape priors but struggle with irregular anatomies or require careful initialization. Hybrid clustering-level-set variants \cite{thakur2019optic} improve accuracy at higher computational cost.
Modern deep learning architectures overcome traditional limitations by automatically learning hierarchical features directly from raw data. However, these deep networks introduce a critical challenge regarding their inherently non-interpretable, "black-box" nature\cite{Singhetal2020}. Unlike classical model-based approaches governed by transparent mathematical constraints-such as explicit objective functions in FCM clustering or energy minimization in active contours\cite{Chenetal2023}---deep networks conceal their decision making pathways within highly complex parameter spaces, hindering clinical trust\cite{Singhetal2020}. By contrast, the traceable nature of traditional frameworks offers distinct practical advantages: every segmentation step-from parameter initialization to contour evolution-is explicitly traceable and mathematically justifiable\cite{Chenetal2023}. This transparency enables developers and clinical engineers to precisely pinpoint systemic errors by reviewing the intermediate computational states, facilitating targeted algorithmic refinement without retraining from scratch. Furthermore, deep learning methods remain constrained by their heavy reliance on extensive annotations and demonstrate poor generalization across varying acquisition protocols\cite{Zhangetal2020}. Consequently, traditional image processing techniques emerge as a uniquely compelling alternative for clinical deployment, where inherent traceability translates directly into enhanced clinical trust and clearer pathways for iterative algorithmic refinement. \\
\begin{figure*}[t]
    \centering
    \includegraphics[width=\linewidth]{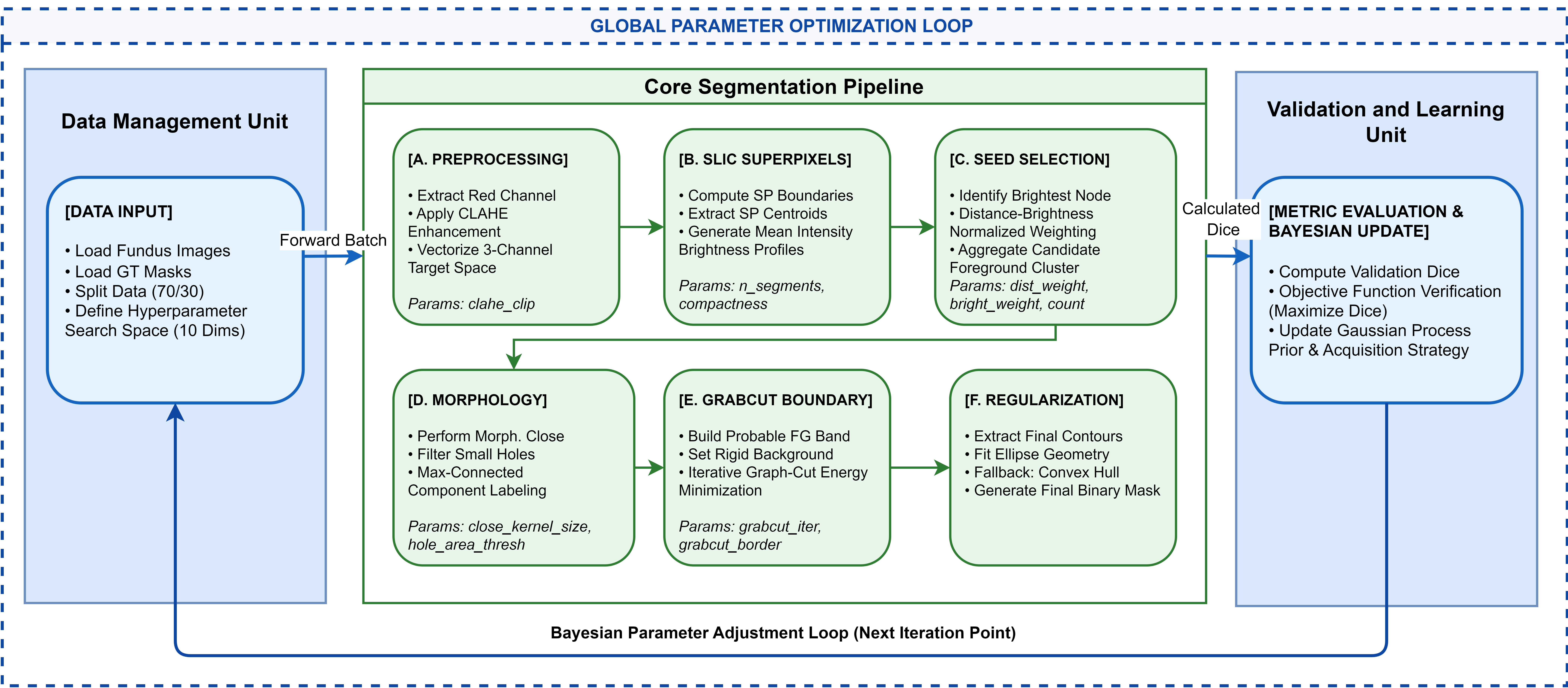}
    \caption{Framework for OD Segmentation and Hyperparameter Optimization}
    \label{fig:pipeline}
\end{figure*}
In this work, we propose a fully trainable, traceable OD segmentation pipeline designed to accurately delineate the OD from fundus images without resorting to deep learning networks. The primary contributions of the paper are summarized as follows:
\begin{enumerate}
    \item We introduce a mathematically transparent OD segmentation pipeline that integrates superpixel-based seed selection and scoring, GrabCut boundary refinement, and elliptical shape regularization, providing an auditable alternative to opaque black-box models.
    \item We formulate hyperparameter tuning as a global objective function solved via Bayesian optimization, eliminating manual parameter tuning while ensuring optimal parameter selection.
    \item We perform an extensive component-wise ablation study and evaluated our approach against state-of-the-art methods, demonstrating that rigorous traditional frameworks can achieve competitive performance without deep learning overhead.
\end{enumerate}
\IEEEpubidadjcol

\section{Methodology}
The proposed OD segmentation aims to generate a binary mask of the OD region using parameterized pipeline that combines superpixel-based foreground seed selection, graph-cut refinement, morphological regularization, and ellipse fitting. To maximize the segmentation performance, we formulate hyperparameter tuning as a global optimization problem solved by Bayesian optimization. The process is designed to be computationally efficient while adapting to variations in image contrast and disc appearance. An overview of the pipeline is given in figure \ref{fig:pipeline}, and each component is detailed mathematically below.

\subsection{Segmentation Pipeline}
Let $I\in\mathbb{R}^{H\times W\times 3}$ denote an RGB image. The red channel $I_R$ is extracted because it typically exhibits the highest contrast between the pale OD and the darker surrounding neuroretinal rim and choroid.

\subsubsection*{Superpixel Decomposition}
To improve the separability of the OD from the background, we apply Contrast Limited Adaptive Histogram Equalization (CLAHE) \cite{Zuiderveld1994CLAHE} to the red channel $I_R$, with a tunable clip limit $\gamma$ that controls the contrast amplification. Adaptive histogram equalization helps to normalize local intensity variations caused by uneven illumination. The enhanced image is converted into a three-channel representation and segmented into superpixels using the Simple Linear Iterative Clustering (SLIC) algorithm \cite{achanta2012slic}.
\begin{equation}
    \mathcal{S}=\text{SLIC}(I_{enh}, K, m, \sigma=1)
\end{equation}
where $\mathcal{S}$ assigns each pixel a superpixel label $s={0, ..., K-1}$, $K$ is the total number of superpixels, $m$ is the compactness, and $\sigma$ is the smoothing scale. SLIC groups perceptually similar pixels while preserving boundaries, which is essential for capturing OD margin.

\subsubsection*{Superpixel Scoring and Selection}
We propose two complementary scoring strategies for superpixel selection: a brightness score $\tilde{b}_s$ that favors high mean intensity in enhanced red channel, and a proximity score $\tilde{d}_s$ that favors spatial closeness to the brightest seed superpixel. This complementary scoring exploits the OD's brightness and spatial cohesion, using the brightest superpixel as a reliable anchor.
For each superpixel $s$, we compute its mean intensity in the enhanced red channel: 
\begin{equation}
    \mu_s = \frac{1}{|\mathcal{P}_s|}\sum_{(x,y)\in \mathcal{P}_s}I_{enh}(x,y)
\end{equation}
where $\mathcal{P}_s$ is the set of pixel coordinates belonging to superpixel $s$. A brightness score $b_s=\mu_s/255$ is computed that normalizes the intensity.
Let $\textbf{c}_s=(\bar{y}_s, \bar{x}_s)$ be the centroid of superpixel $s$ (row, column indices). The superpixel with the highest brightness is chosen as the seed: 
\begin{equation}
    s^*=arg\max_s(b_s), \quad \textbf{c}^*=\textbf{c}_s^*
\end{equation}
For all superpixels, we compute the normalized Euclidean distance from the seed centroid:
\begin{equation}
    \tilde{d}_s=\frac{||\textbf{c}_s-\textbf{c}^*||_2}{\max_{s'}||\textbf{c}_{s'}-\textbf{c}^*||_2+\epsilon}
\end{equation}
and the normalized inverse brightness (to favor bright regions):
\begin{equation}
    \tilde{b}_s=\frac{b_s}{max_{s'}b_{s'}+\epsilon}
\end{equation}
A combined score $q_s$ that balances proximity to the seed and high intensity is defined as:
\begin{equation}
    q_s=w_d\tilde{d}_s+w_b(1-\tilde{b}_s)
\end{equation}
where $w_d, w_b \in [0,1]$ (tunable) control the influence of distance and brightness, respectively. The top $k$ superpixels with the smallest $q_s$ (i.e, closest and brightest) are selected
. The seed superpixel is forcibly included if not already among the selected ones. The initial OD mask $M_{init}$ is the union of all pixels belonging to the selected superpixels.

\subsubsection*{Morphological Refinement}
The initial mask may contain small holes or spurious projections due to vessels. We apply morphological closing with an elliptical structuring element (SE) of size $\kappa$ (tunable):
\begin{equation}
    M_{close}=(M_{init}\oplus \text{SE}) \ominus \text{SE}
\end{equation}
where $\oplus$ and $\ominus$ denote dilation and erosion. This fills narrow gaps and smooths boundary.
Small holes (areas below a threshold $\tau_{hole}$ tunable from 50 to 500 pixels) are then removed using morphological hole filling:
\begin{equation}
    M_{filled}=\text{remove\_small\_holes}(M_{close},\tau_{hole})
\end{equation}
The resulting mask may still contain multiple disconnected components. To retain only the component that most likely corresponds to the OD, we identify the connected component that contains the seed superpixel $s^*$. Let $\{C_1, ..., C_m\}$ be the connected components of $M_{filled}$. The final localization mask is: 
\begin{equation}
    M_{loc}=C_j \qquad \text{where} \quad C_j\cap\mathcal{P}_{s^*} \neq \phi
\end{equation}
If the seed superpixel falls outside all components, we instead select the largest component by area. 
This localization mask $M_{loc}$ serves as a coarse OD region. It is subsequently passed to boundary refinement stage for precise delineation. 

\subsubsection*{GrabCut Boundary Refinement}
To improve adherence to the true disc boundaries, we refine $M_{loc}$ using GrabCut\cite{Rother2004GrabCut}, an iterative graph-cut algorithm that minimizes an energy functional. A band of width $\beta$ (tunable) around the mask is treated as "probable background", while the inner region is "probable foreground". Formally, we define: 
\begin{equation}
    \Omega_{fg}=\{(x,y):M_{loc}=1\}
\end{equation}
\begin{equation}
    \Omega_{bg}=\{(x,y):M_{loc}=0\text{ and }dist((x,y), \Omega_{fg})>\beta)\}
\end{equation}
\begin{equation}
    \Omega_{unk}=\{(x,y):M_{loc}=0 \text{ and } 0<dist((x,y),\Omega{fg})\leq\beta\}
\end{equation}
These sets define the GrabCut trimap. GrabCut iteratively minimizes a standard energy function---combining Gaussian-mixture color likelihoods and contrast-sensitive pairwise smoothness---via min-cut/max-flow over $T_{gc}$ iterations, yielding a refined mask $M_{gc}$

\subsubsection*{Ellipse Regularization}
The OD is approximately elliptical. To enforce a smooth, convex shape, we fit least-squares ellipse to the contour of $M_{gc}$. Let $\partial M_{gc}$ be the set of boundary points. An ellipse is defined by its center $(c_x, c_y)$, major axis length $a$, and minor axis length $b$, and orientation $\theta$.
\begin{multline}
    \min_{c_x, c_y, a, b, \theta}\sum_{(x,y)\in\partial M_{gc}}\Bigg(\frac{((x-c_x)cos\theta+(y-c_y)sin\theta)^2}{a^2}\\
    +\frac{(-(x-c_x)sin\theta+(y-c_y)cos\theta)^2}{b^2}-1\Bigg)^2
\end{multline}
If the ellipse fitting fails, we fall back to the convex hull of $\partial M_{gc}$. The final segmentation mask $M_{final}$ is obtained by rasterizing the fitted ellipse.

The complete sequence of operations---from superpixel decomposition to ellipse regularization---is consolidated in algorithm \ref{algo:main}.
\begin{algorithm}[p]
\caption{Guided SLIC with Brightness and Proximity Prior for OD Segmentation}
\label{algo:main}
\begin{algorithmic}[1]
\Require RGB fundus image $\mathbf{I}\in \mathbb{R}^{H\times W\times 3}$
\Ensure Binary OD mask $\textbf{M}_{final}$

\State \textbf{Step 1: Superpixel Decomposition}
\State $\mathbf{I}_R \gets \mathbf{I}(:,:,0)$ \Comment{Extract red channel}
\State $\mathbf{I}_{enh} \gets \text{CLAHE}(\mathbf{I}_R, \gamma)$ \Comment{Apply CLAHE}
\State $\mathcal{S} \gets \text{SLIC}(\mathbf{I}_{enh},K,m)$

\State \textbf{Step 2: Superpixel Scoring and Selection}
\For {each superpixel $s_i \in \mathcal{S}$}
\State $\mu_i \gets \frac{1}{|s_i|}\sum_{(x,y)\in s_i} \mathbf{I}_{enh}(x,y)$
\State $b_i \gets \frac{\mu_i}{255}$
\EndFor
\State $s^* \gets \arg\max_i b_i$ \Comment{Identify seed superpixel}
\State $\mathbf{c}_i \gets \frac{1}{|s_i|}\sum_{(x,y) \in s_i}(x,y)$ \Comment{Compute centroid}
\State $d_i \gets ||\mathbf{c}_i-\mathbf{c}_{s^*}||_2$ \Comment{Compute distances to seed}
\State $\tilde{d}_i \gets \frac{d_i}{max_j d_j+\epsilon}, \tilde{b}_i \gets \frac{b_i}{max_j b_j+\epsilon}$
\State $q_i \gets w_d \cdot \tilde{d}_i + w_b \cdot (1-\tilde{b}_i)$
\State $\mathcal{S}_{set} \gets \text{Top}_k({\mathcal{S},q_i})$
\State $\textbf{M}_{init} \gets \textbf{0}$
\For{each $s_i \in \mathcal{S}_{set}$}
    \State $\textbf{M}_{init}(x,y) \in 1, \forall (x,y) \in s_i$
\EndFor
\State \textbf{Step 3: Morphological Refinement}
\State $\text{SE}_{\kappa} = \left\{ (x,y) \in \mathbb{Z}^2 \mid x^2 + y^2 \leq \kappa^2 \right\}$
\State $\textbf{M}_{close} \gets (\textbf{M}_{init} \oplus \text{SE}_{\kappa}) \ominus \text{SE}_{\kappa}$
\State $\textbf{M}_{filled} \gets \text{remove\_small\_holes}(\textbf{M}_{close}, \tau_{hole})$
\State $\{\mathcal{C}_j\} \gets \text{connected\_components}(\mathbf{M}_{filled})$
\State $\mathcal{C}^* \gets \{ \mathcal{C}_j \mid \mathcal{C}_j \cap \mathcal{P}_{s^*} \neq \emptyset \}$ \Comment{Find component(s) overlapping the seed}
\If{$|\mathcal{C}^*| > 1$}
    \State $\mathcal{C}_{target} \gets \arg\max_{\mathcal{C} \in \mathcal{C}^*} |\mathcal{C}|$ \Comment{Pick largest among those containing the seed}
\Else
    \State $\mathcal{C}_{target} \gets \mathcal{C}^*$
\EndIf
\State $\mathbf{M}_{loc} \gets \mathbbm{1}[(x,y) \in \mathcal{C}_{target}]$
\State \textbf{Step 4: GrabCut Boundary Refinement}
\State $\Omega_{fg}=\{(x,y)\mid M_{loc}(x,y)=1\}$
\State $\text{SE}_{B_g}\gets \{(x,y)\in\mathbb{Z}^2 \mid x^2+y^2\leq B^2_g\}$
\State $\textbf{M}_{dilate}\gets\textbf{M}_{loc}$
\State $\Omega_{unk}=\{(x,y)\mid \textbf{M}_{dil}(x,y)=1 \text{ and } \textbf{M}_{loc}(x,y)=0\}$
\State $\Omega_{bg}=\{(x,y)\mid\textbf{M}_{dil}(x,y)=0\}$
\State $E(f)=\sum_{p} D_p(f_p) + \lambda \sum_{(p,q)\in\mathcal{N}} V_{pq}(f_p, f_q)$
\State $\hat{f} = \arg\min_{f \in \{0,1\}^{H \times W}}E(f)$
\State $\textbf{M}_{gc}\gets\mathbbm{1}[\hat{f}=1]$
\State \textbf{Step 5: Ellipse Regularization}
\State $\Gamma \gets \partial\textbf{M}_{gc}$ \Comment{Extract contours}
\State $\mathcal{E} \gets \text{fit\_ellipse\_ls}(\Gamma)$ \Comment{Fit ellipse $\mathcal{E}$ via least squares}
\If{$\mathcal{E}$ fits}
\State $\textbf{M}_{final} \gets \text{Rasterize}(\mathcal{E})$
\Else
\State $\textbf{M}_{final} \gets \text{ConvexHull}(\partial \textbf{M}_{gc})$
\EndIf

\Return $\textbf{M}_{final}$
\end{algorithmic}
\end{algorithm}

\subsection{Hyperparameter Optimization}
\label{subsec:hyperparameter}
Manual tuning is impractical and dataset-specific. We therefore formulate hyperparameter optimization as a minimization problem: 
\begin{equation}
\begin{split}
    \theta^*&=arg \min_{\theta}\mathcal{L}(\theta)\\
    \quad \mathcal{L}(\theta)&=-\frac{1}{N_{train}}\sum^{N_{train}}_{i=1}\text{Dice}(f(I_i;\theta),G_i),
\end{split}
\end{equation}
where $f(I_i;\theta)$ is the predicted binary mask for image $I_i$ using parameter configuration $\theta$, $G_i$ is the ground truth, and Dice $(\cdot,\cdot)$ represents the Dice coefficient. The objective $\mathcal{L}(\theta) \in [-1,0]$ (negative Dice) is minimized. The parameter space $\Theta$ is defined as a Cartesian product of independent intervals: integer parameters are sampled from discrete domains $[L_j,U_j]_\mathbb{Z}$ and real parameters from continuous domains $[L_j,U_j]\subset\mathbb{R}$.
To effectively minimize $\mathcal{L}(\theta)$, we adopt Bayesian optimization (BO)\cite{shahriari2016loop}, which is particularly suited for low-to-moderate dimensional problems where each objective evaluation is computationally expensive. 
We use a standard GP surrogate (Mat\`ern 5/2 kernel) with Expected Improvement (EI) acquisition. 
After $t$ evaluations, $\{\theta_i,y_i\}_{i=1}^t$ (where $y_i=\mathcal{L}(\theta_i)$), the posterior predictive distribution at a new point $\theta$ is Gaussian with mean $\mu_t(\theta)$ and variance $\sigma^2_t(\theta)$.\\
The next evaluation point, $\theta_{t+1}$ is selected using the EI function.
\begin{equation}
    \theta_{t+1}=arg \max_{\theta \in \Theta}\text{EI}(\theta)
\end{equation}
After a predefined number of iterations $K_{BO}$, the parameter configuration $\theta^*$ with the lowest observed objective value is returned. The moderate iteration budget is a pragmatic choice driven by the limited validation set size, where extended BO runs would risk overfitting the surrogate to the validation split without improving true held-out generalization. 
The automated tuning enables the pipeline to adapt without manual intervention, ensuring reproducibility and optimal performance. Crucially, this optimization is deployed only during offline model selection the final inference pipeline (Algorithm \ref{algo:main}) operates with fixed parameters and is entirely deterministic, ensuring that the runtime segmentation remains fully mathematically traceable without any opaque optimization overhead.

\section{Experiments}
\subsection{Dataset}
We evaluated our method on the publicly available Drishti-GS\cite{Sivaswamy2014DrishtiGS} fundus image dataset. This dataset comprises 101 color retinal fundus photographs captured at a resolution of $2896 \times 1944$ pixels, providing expert-annotated pixel-level masks for the background, OD, and optic cup (OC). For our evaluation, we adhered to the official data split of 50 training and 51 testing images. To optimize hyperparameters, we further partitioned the official training set into a training subset (70\%, 35 images) and a validation subset (30\%, 15 images) to monitor generalization during development, while strictly holding out the official test set for final performance assessment. Following standard practice, we binarized the ground-truth OD masks using a threshold of 127 to obtain binary segmentation targets.
\subsection{Implementation details}
The complete pipeline is implemented as a modular framework comprising the core segmentation routine and a hyperparameter optimizer. The core routine processes the images at native resolution, with all morphological operations using elliptical structuring elements. The proposed pipeline contains 10 tunable parameters, listed in Table \ref{tab:tunable}, with integer parameters sampled uniformly from discrete sets and real parameters from continuous intervals. 
\begin{table}[h]
    \centering
    \caption{Tunable parameters for OD segmentation pipeline (* refers to the parameters $\in \mathbb{Z}$, $\dagger$ refers to the parameters $\in\mathbb{R})$}
    \label{tab:tunable}
    \renewcommand{\arraystretch}{1.2}
    \begin{tabular}{|l|c|c|} \hline
        \textbf{Parameter} & \textbf{Symbol} & \textbf{Range} \\ \hline
        \texttt{clahe\_clip}$^\dagger$ & $\gamma$ & [1.0, 4.0] \\ \hline
        \texttt{n\_segments}$^*$ & $K$ & [300, 700] \\ \hline
        \texttt{compactness}$^\dagger$ & $m$ & [1, 30] \\ \hline
        \texttt{dist\_weight}$^\dagger$ & $w_d$ & [0.3, 0.9] \\ \hline
        \texttt{bright\_weight}$^\dagger$ & $w_b$ & [0.1, 0.7] \\ \hline
        \texttt{superpixel\_count}$^*$ & $k$ & [7, 15] \\ \hline
        \texttt{close\_kernel\_size}$^*$ & $\kappa$ & [3, 15] \\ \hline
        \texttt{hole\_area\_thresh}$^*$ & $\tau_{hole}$ & [50, 500] \\ \hline
        \texttt{grabcut\_iter}$^*$ & $T_{gc}$ & [2, 8] \\ \hline
        \texttt{grabcut\_border}$^*$ & $\beta$ & [10, 20] \\ \hline
    \end{tabular}
\end{table}
Hyperparameter optimization follows the Bayesian strategy, implemented using a Gaussian process surrogate with a Mat\`ern $5/2$ kernel and Expected Improvement acquisition. The optimization executes 10 iterations: 2 initial random points to seed the surrogates, followed by 8 GP-guided evaluations. A fixed random seed (42) ensures reproducible $70/30$ training-validation splits. The objective function minimizes the negative mean Dice coefficient over the training subset, while validation Dice is monitored at each iteration; the configuration achieving the highest validation performance is retained as final model. The self-contained implementation executes on CPU, requiring no external pre-trained weights.

\subsection{Evaluation Metrics}
To quantitatively evaluate the segmentation performance, we adopt Dice similarity coefficient, sensitivity (Sens.), specificity (Spec.), and precision (Prec.) derived from the pixel-wise confusion matrix.
\begin{align}
    \text{Dice} &= \frac{2TP}{2TP + FP + FN}, & 
    \text{Sens.} &= \frac{TP}{TP + FN}, & \\
    \text{Spec.} &= \frac{TN}{TN + FP}, & 
    \text{Prec.} &= \frac{TP}{TP + FP}
\end{align}
where \(TP\), \(TN\), \(FP\), and \(FN\) denote true positives, true negatives, false positives, and false negatives, respectively. 

\subsection{Experiments and Results}
The component-wise study in Table \ref{tab:ablation} isolates each module's contribution. The full pipeline achieves a Dice coefficient of 0.9536. The exclusion of CLAHE drops the Dice to 0.8956 and the precision to 0.8879, which confirms its essential role in contrast normalization. The removal of proximity weighting severely harms sensitivity (0.8551), which proves that spatial cues are critical for recovering the complete disc. Brightness weighting has a milder effect, as its omission yields a Dice of 0.9456. \\
\begin{table}[h]
    \centering
    \caption{Component-wise Analysis for OD Segmentation}
    \renewcommand{\arraystretch}{1.3}
    \begin{tabular}{|l|c|c|c|c|} \hline
        \textbf{Configuration} & \textbf{Dice} & \textbf{Sens.} & \textbf{Spec.} & \textbf{Prec.} \\ \hline
        \textbf{Full Pipeline} & \textbf{0.9536} & \textbf{0.9255} & 0.9995 & 0.9811\\ \hline
        \multicolumn{5}{|l|}{\textit{Preprocessing}} \\ \hline
        w/o CLAHE Enhancement & 0.8956 & 0.9077 & 0.9956 & 0.8879 \\ \hline
        \multicolumn{5}{|l|}{\textit{Segmentation}} \\ \hline
        w/o Proximity Weighting & 0.9048 & 0.8551 & 0.9995 & 0.9826 \\ 
        w/o Brightness Weighting & 0.9456 & 0.9279 & 0.999 & 0.9683 \\ \hline
        \multicolumn{5}{|l|}{\textit{Refinement}} \\ \hline
        w/o Morphology Refinement & 0.9314 & 0.9076 & 0.9989 & 0.9639 \\ 
        w/o GrabCut Refinement & 0.8742 & 0.8689 & 0.9967 & 0.8922 \\ 
        w/o Ellipse Regularization & 0.9384 & 0.8962 & 0.9996 & 0.9869 \\ \hline
        \multicolumn{5}{|l|}{\textit{Optimization}} \\ \hline
        w/o Bayesian Optimization & 0.7105 & 0.5543 & \textbf{1.0000} & \textbf{0.9993} \\ \hline 
    \end{tabular}
    \label{tab:ablation}
\end{table}
\begin{figure}[htbp]
    \centering
    \begin{subfigure}[b]{0.155\textwidth}
        \centering
        \includegraphics[width=\textwidth]{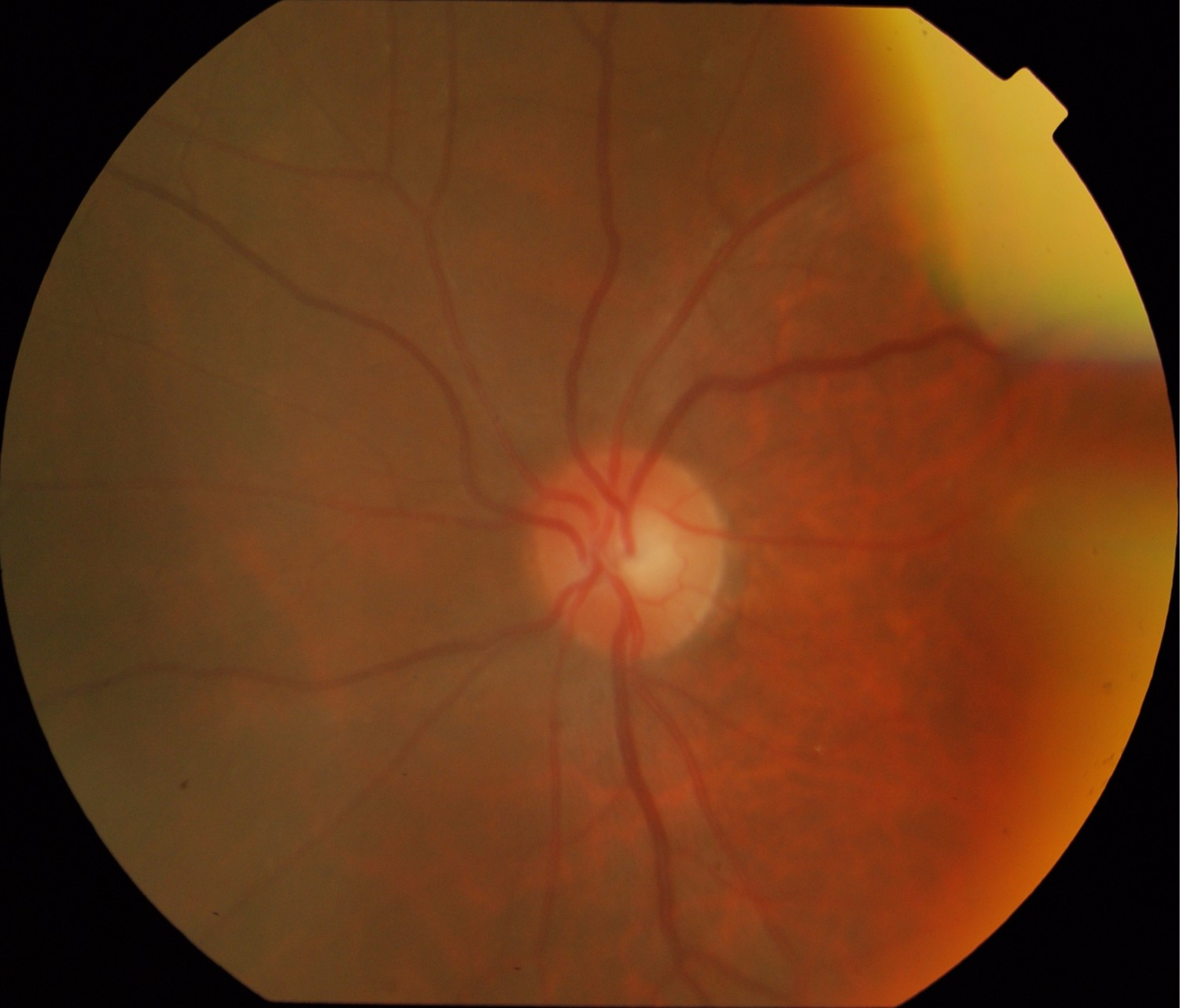}
        \caption{}
        \label{fig:input_easy}
    \end{subfigure}
    \hfill
    \begin{subfigure}[b]{0.155\textwidth}
        \centering
        \includegraphics[width=\textwidth]{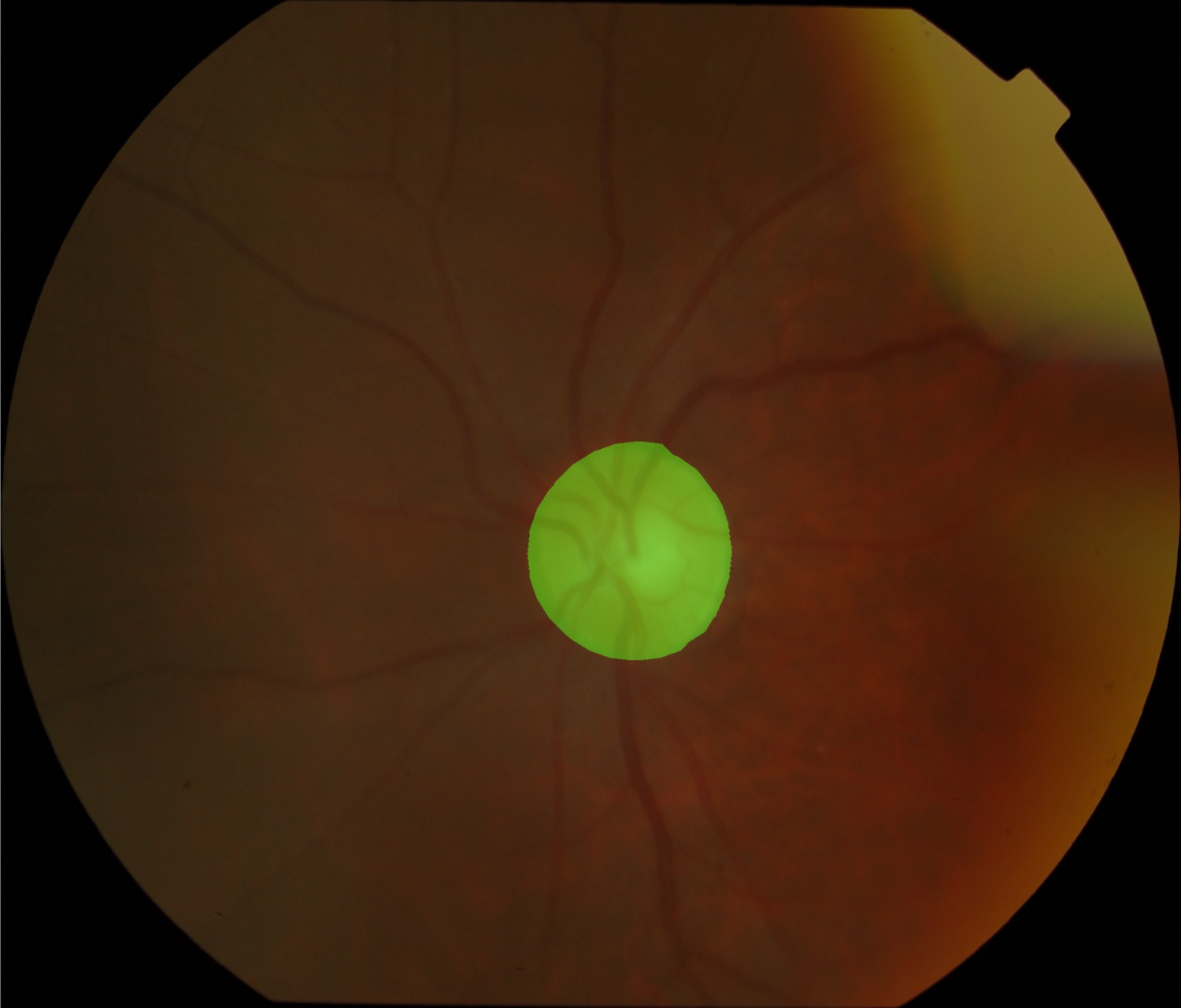}
        \caption{}
        \label{fig:gt_easy}
    \end{subfigure}
    \hfill
    \begin{subfigure}[b]{0.155\textwidth}
        \centering
        \includegraphics[width=\textwidth]{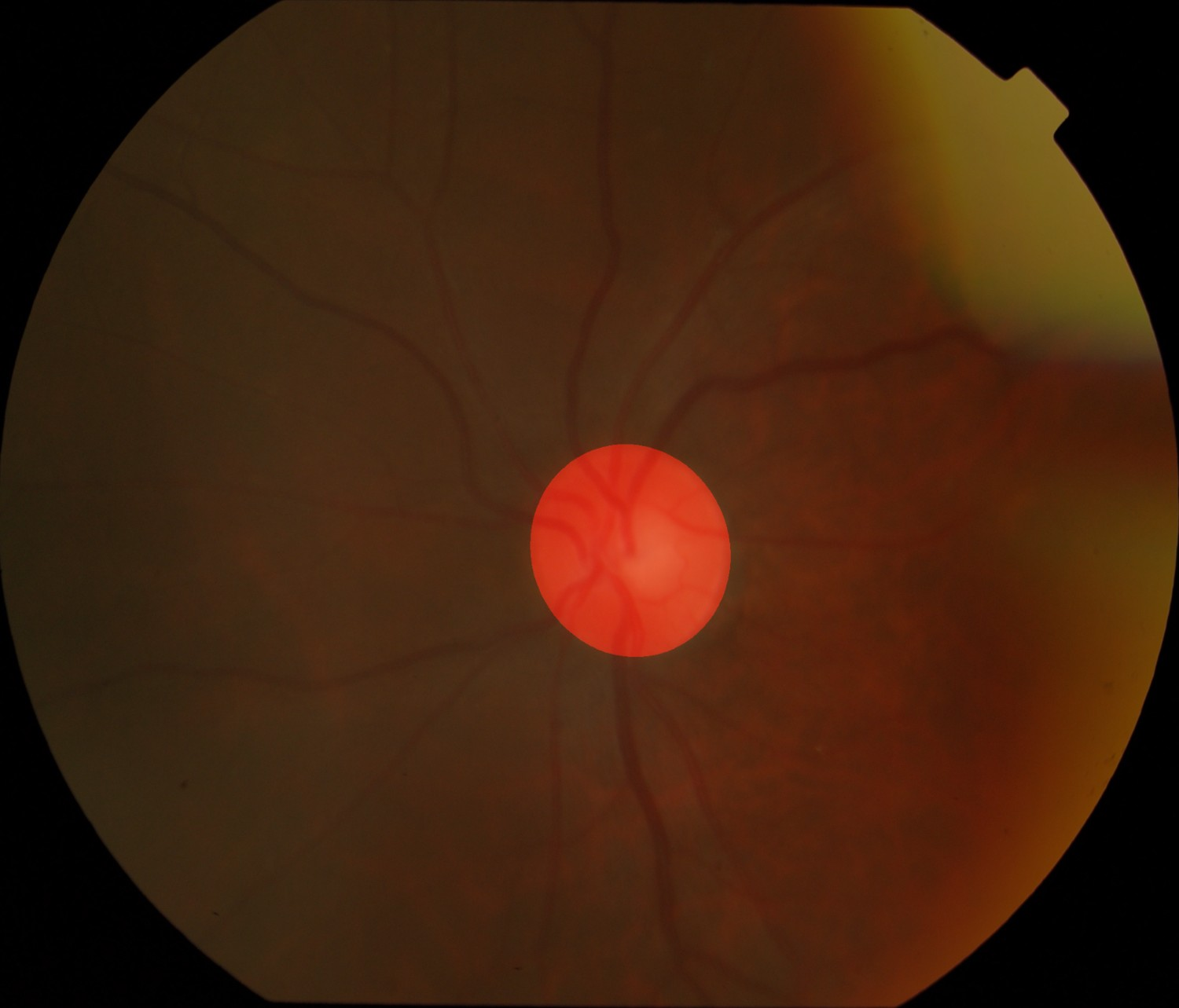}
        \caption{}
        \label{fig:pred_easy}
    \end{subfigure}
    \begin{subfigure}[b]{0.155\textwidth}
        \centering
        \includegraphics[width=\textwidth]{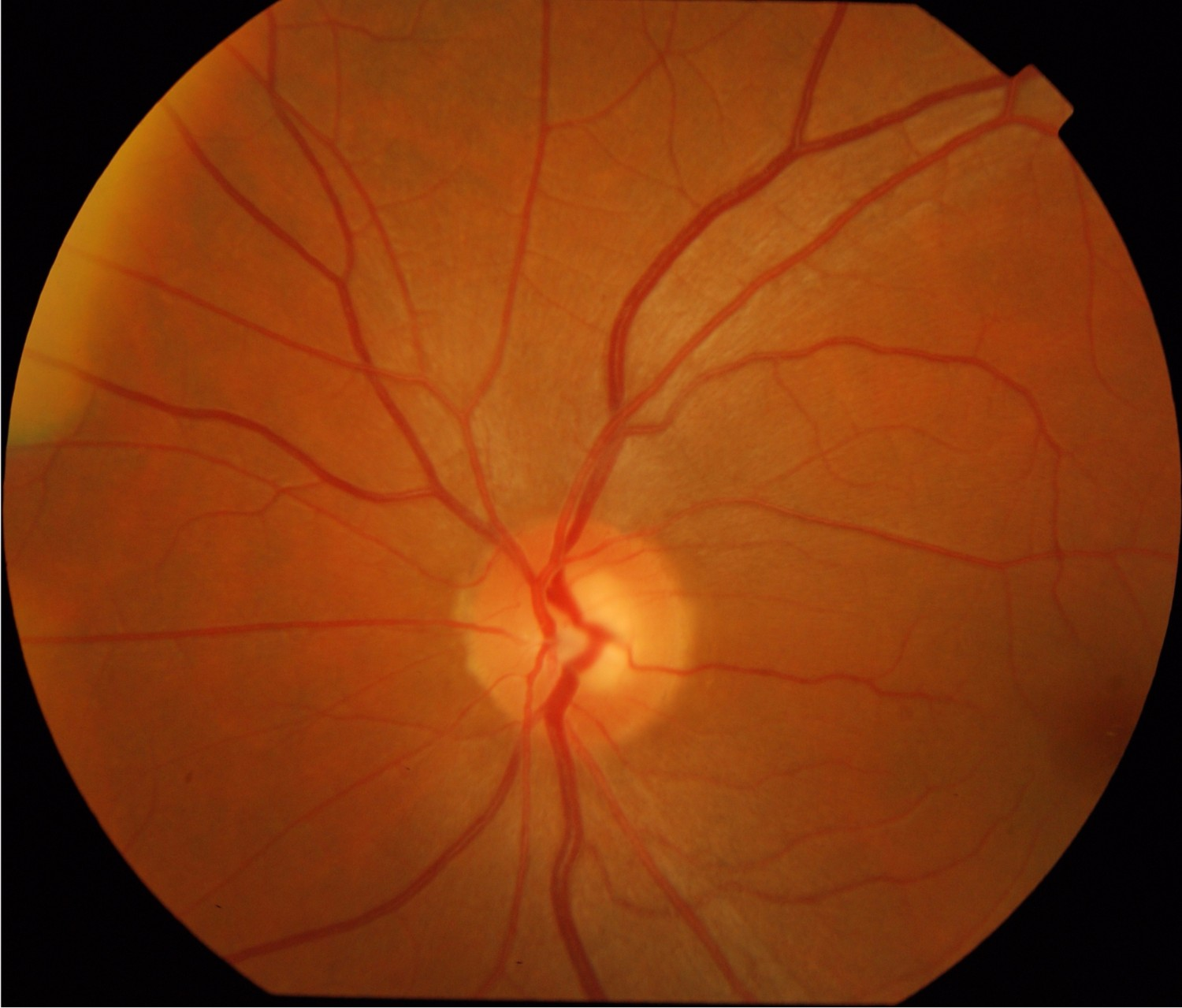}
        \caption{}
        \label{fig:input_hard}
    \end{subfigure}
    \hfill
    \begin{subfigure}[b]{0.155\textwidth}
        \centering
        \includegraphics[width=\textwidth]{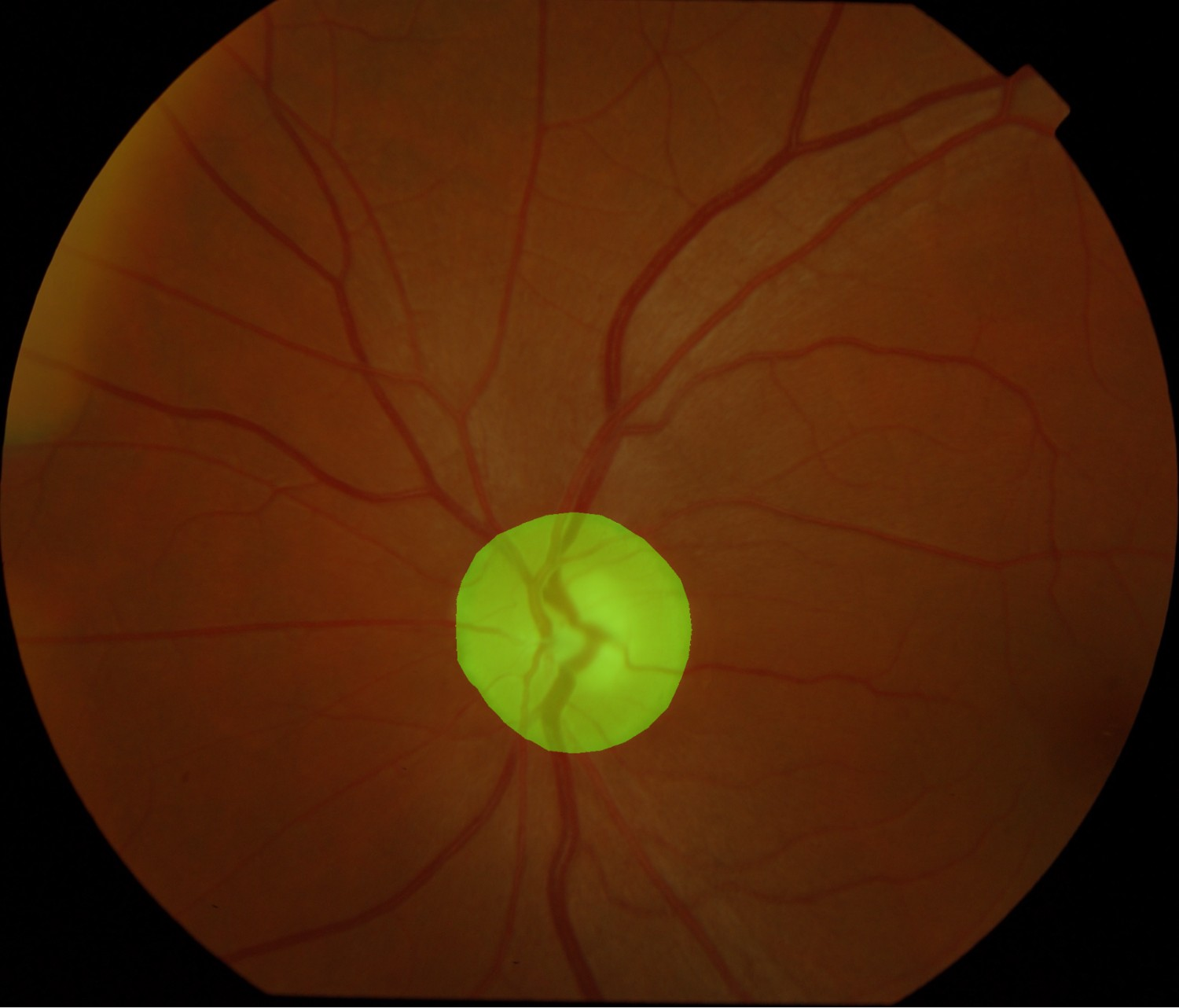}
        \caption{}
        \label{fig:gt_hard}
    \end{subfigure}
    \hfill
    \begin{subfigure}[b]{0.155\textwidth}
        \centering
        \includegraphics[width=\textwidth]{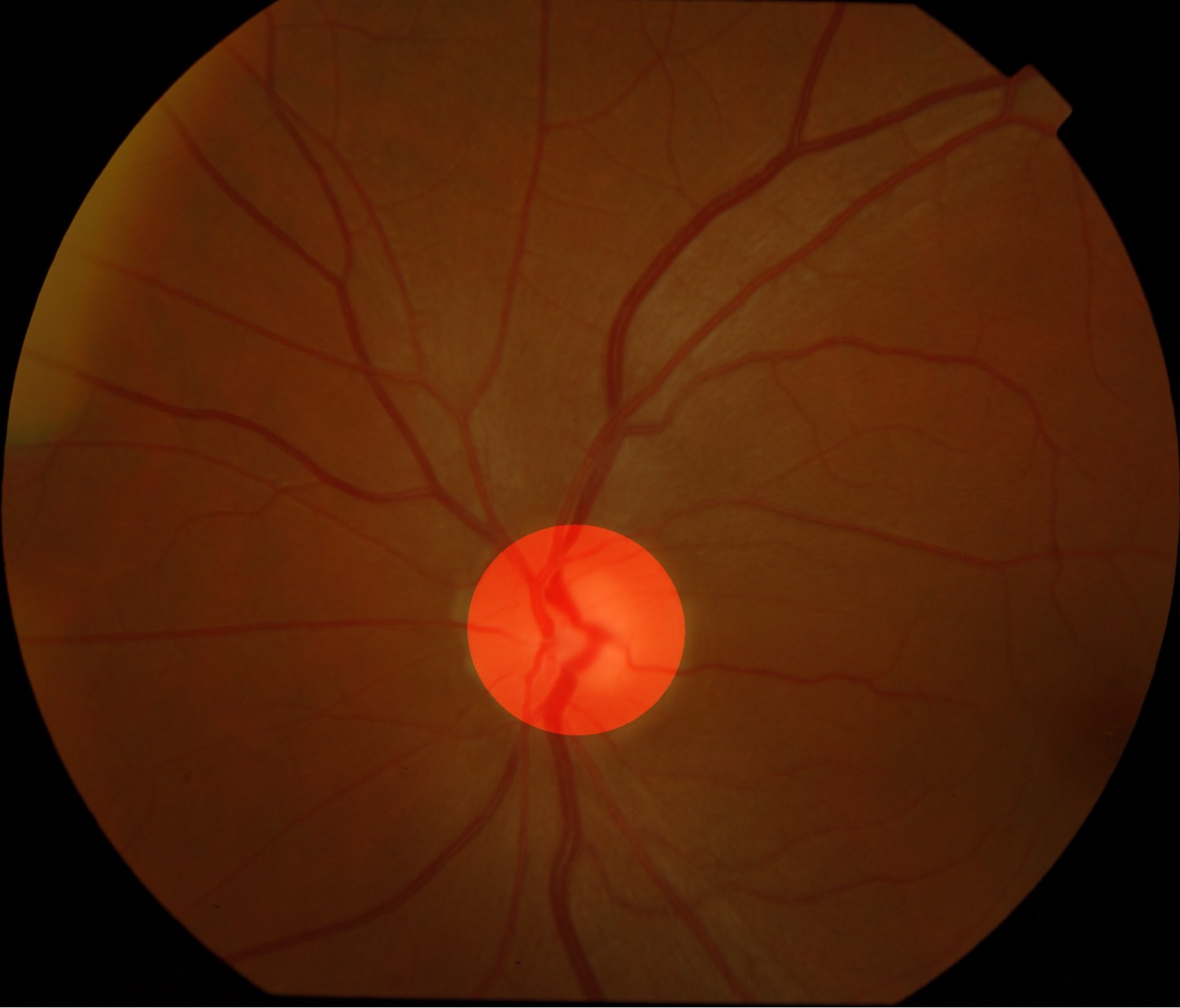}
        \caption{}
        \label{fig:pred_hard}
    \end{subfigure}
    \begin{subfigure}[b]{0.155\textwidth}
        \centering
        \includegraphics[width=\textwidth]{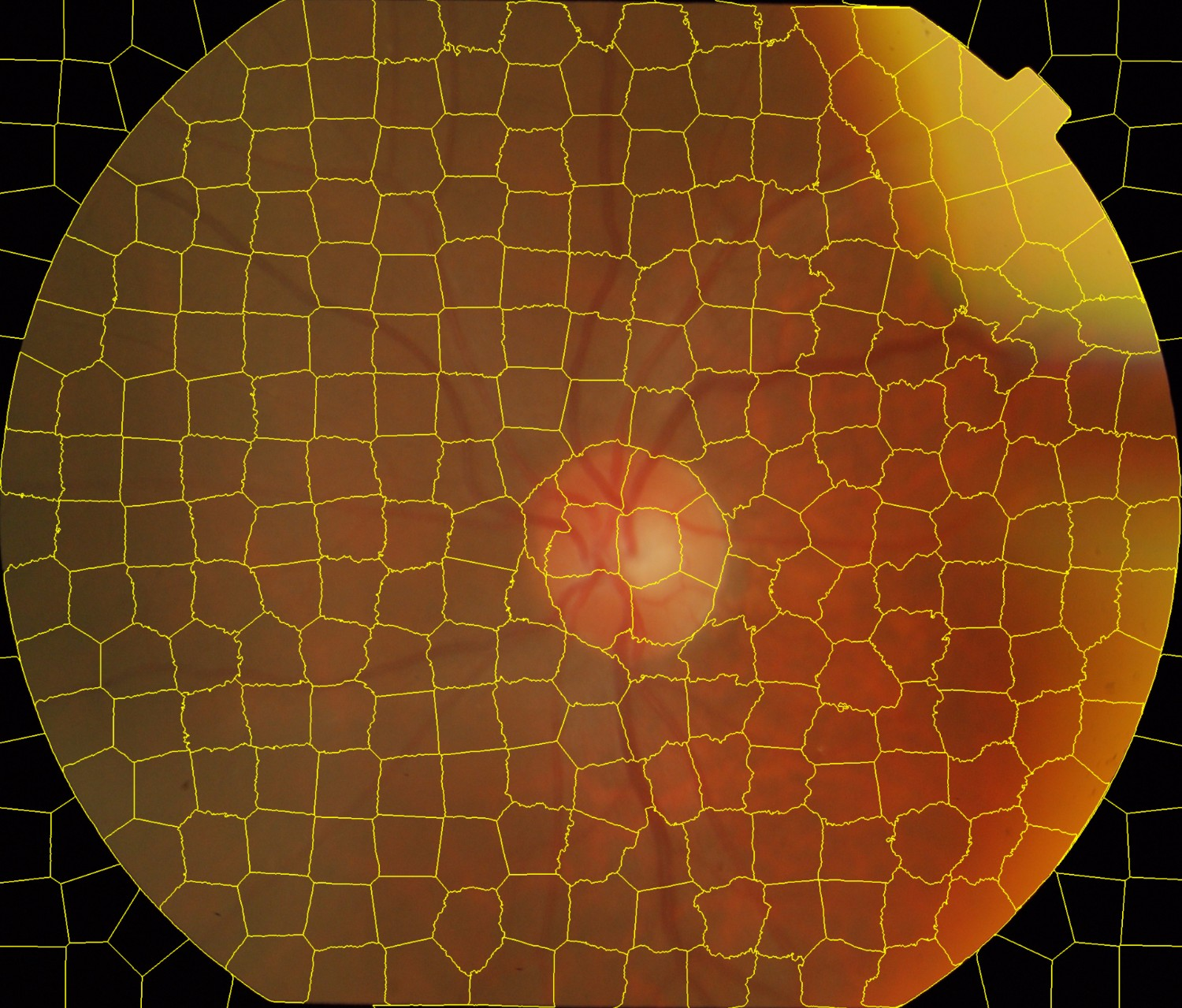}
        \caption{}
        \label{fig:slic}
    \end{subfigure}
    \hfill
    \begin{subfigure}[b]{0.155\textwidth}
        \centering
        \includegraphics[width=\textwidth]{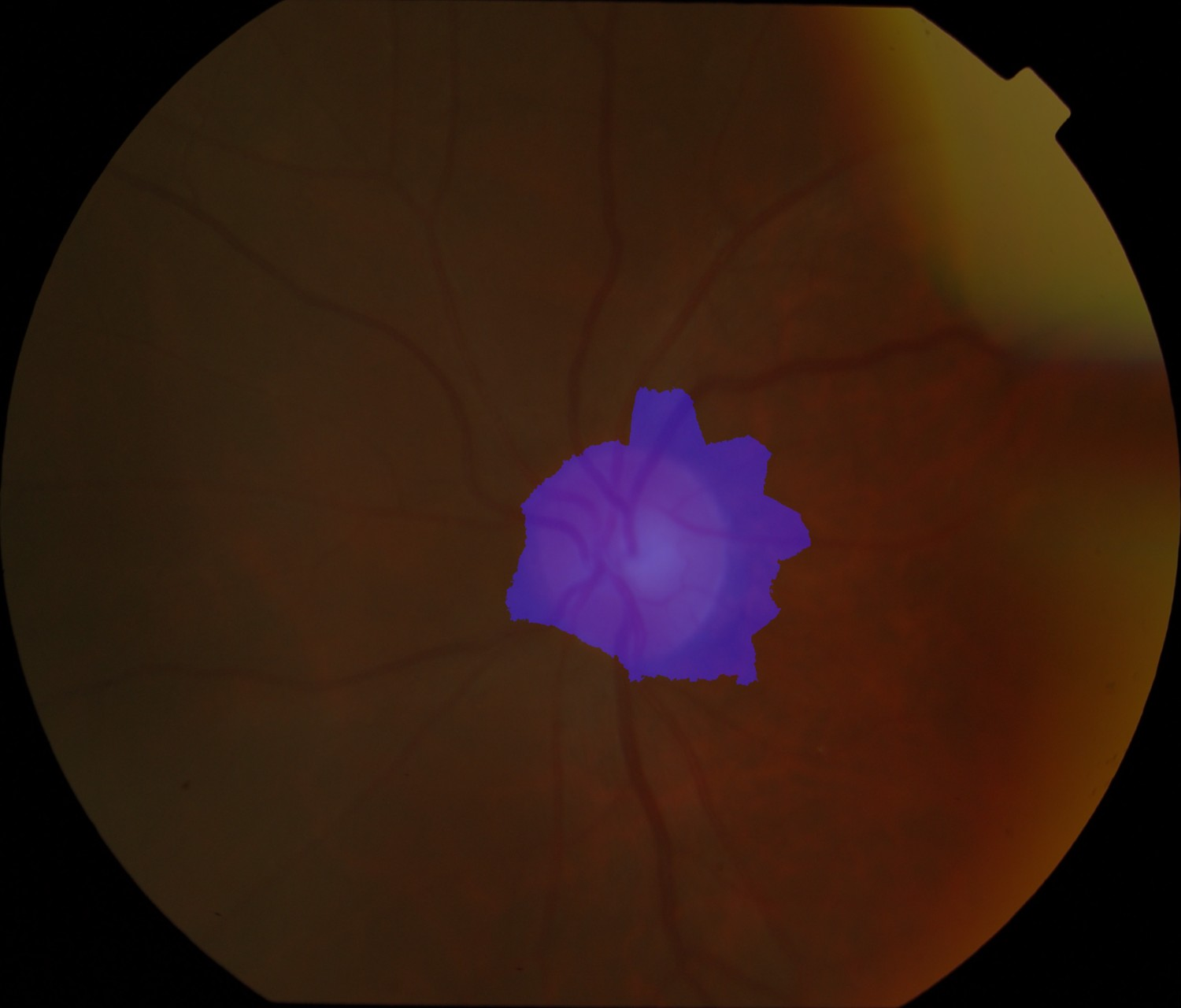}
        \caption{}
        \label{fig:init_mask}
    \end{subfigure}
    \hfill
    \begin{subfigure}[b]{0.155\textwidth}
        \centering
        \includegraphics[width=\textwidth]{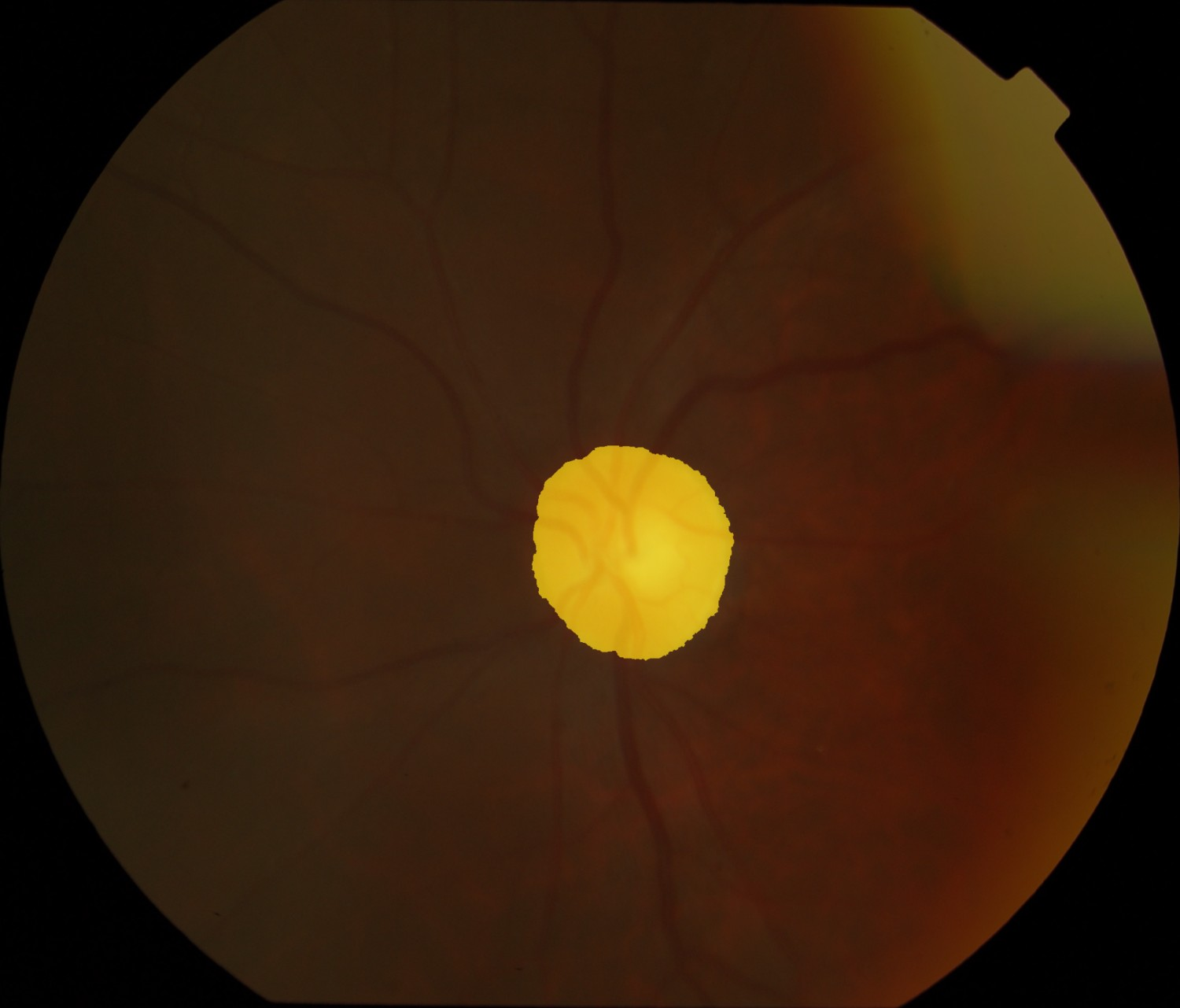}
        \caption{}
        \label{fig:gc_ellipse}
    \end{subfigure}
    \caption{Qualitative segmentation results on the Drishti-GS dataset. (a), (d) Input fundus images; (b), (e) Corresponding ground-truth masks; (c), (f) Corresponding predictions with Dice scores of 0.972 and 0.897, respectively. The bottom row depicts the intermediate pipeline outputs: (g) SLIC superpixel decomposition, (h) initial coarse mask, and (i) final refined mask after GrabCut and elliptical regularization.}
    \label{fig:qualitative_grid}
\end{figure}
Among the refinement stages, the omission of GrabCut causes the largest degradation, with a Dice of 0.8742 and a precision of 0.8922, which underscores its necessity for boundary fidelity. Morphological closing provides moderate gains by filling vessel-induced gaps, as its absence results in a Dice of 0.9314. Ellipse regularization slightly reduces sensitivity to 0.8962 but improves precision to 0.9869, which reflects the expected trade-off between smooth anatomical priors and irregular shape preservation. Finally, disabling BO collapses performance to a Dice of 0.7105 and a sensitivity of 0.5543, which demonstrates that manual defaults fail entirely. This result validates BO as an effective navigator of the coupled parameter space.
The quantitative results are further corroborated by  qualitative assessment using representative fundus images, as illustrated in figure \ref{fig:qualitative_grid}. In figures \ref{fig:input_easy}, \ref{fig:gt_easy}, and \ref{fig:pred_easy}, which represents an optimal scenario with distinct physiological boundaries, the algorithm adhere precisely to the disc margins, leading to a highly accurate prediction that achieves a Dice score of 97.2\%. Conversely, figures \ref{fig:input_hard}, \ref{fig:gt_hard}, and \ref{fig:pred_hard} presents a challenging scenario characterized by low boundary contrast. Despite the structural irregularities and fragmented edges, our framework delivers an anatomically plausible prediction that maintains a robust Dice score of 89.7\%. The progression across figures \ref{fig:slic}, \ref{fig:init_mask}, \ref{fig:gc_ellipse} demonstrates that while the SLIC and coarse masking phases efficiently isolate the region of interest, the integration of GrabCut and ellipse fitting is vital for mitigating vessel-induced occlusions and ensuring robust boundary fidelity scores across varying clinical morphologies.\\  
\subsubsection*{Comparison with State-of-the-Art Methods}
\begin{table}[t]
    \centering
    \caption{Comparison of our proposed framework with the existing state-of-the-art traditional methods}
    \renewcommand{\arraystretch}{1.3}
    \begin{tabular}{|l|c|c|c|c|} \hline
        \textbf{Method} & \textbf{Dice} & \textbf{Sens.} & \textbf{Spec.} & \textbf{Prec.} \\ \hline
        \multicolumn{5}{|l|}{\textit{Traditional methods}} \\ \hline
        Adams et al.\cite{adams1994seeded} & 0.88 & 0.83 & 0.88 & 0.87 \\
        Adhikari et al.\cite{adhikari2015spatial} & 0.89 & 0.87 & 0.91 & 0.91 \\ 
        Elazab et al.\cite{elazab2015segmentation} & 0.91 & 0.87 & 0.91 & 0.91 \\
        Thakur et al.\cite{thakur2019optic} & 0.92 & 0.89 & 0.92 & 0.92 \\ 
        Haleem et al.\cite{haleem2018novel} & 0.95 & -- & -- & -- \\ \hline
        \multicolumn{5}{|l|}{\textit{Deep Learning Methods}} \\ \hline
        DeepLab v3+\cite{chen2017deeplab} & 0.92 & -- & -- & -- \\ 
        FCN\cite{long2015fully} & 0.95 & -- & -- & -- \\ 
        U-Net\cite{ronneberger2015u} & \textbf{0.96} & \textbf{0.97} & 0.99 & 0.95 \\ \hline
        \textbf{Our Approach} & 0.95 & 0.93 & \textbf{0.99} & \textbf{0.98} \\ \hline
    \end{tabular}
    \label{tab:sota}
\end{table}
A quantitative comparison of the proposed framework against existing state-of-the-art OD segmentation techniques is summarized in Table \ref{tab:sota}. The benchmarked competitors encompass both established traditional paradigms (e.g., region-growing \cite{adams1994seeded}, clustering-based \cite{elazab2015segmentation,thakur2019optic}, and deformable \cite{haleem2018novel} models) and recent deep learning architectures (DeepLab v3+ \cite{chen2017deeplab}, FCN \cite{long2015fully}, and U-Net \cite{ronneberger2015u}). Compared with traditional techniques, our approach attains a Dice of 0.95, matching the highest reported value, while consistently outperforming all of them on sensitivity (0.93), specificity (0.99), and precision (0.98). Relative to deep learning baselines, our method ties with FCN (0.95) and closely follows U-Net (0.96), yet achieves superior precision (0.98 vs. 0.95) and comparable specificity. Collectively, these results demonstrate the proposed framework's effectiveness in delivering accurate, robust, and traceable OD delineation.

\section{Conclusion}

We have presented a traceable OD segmentation pipeline that integrates SLIC superpixel selection, morphological refinement, GrabCut delineation, and elliptical regularization, with hyperparameters tuned via Bayesian optimization. On the Drishti-GS dataset, our method achieves a Dice of 0.9536-competitive with the best traditional approaches-while offering full algorithmic traceability for regulatory review. The component-wise ablation study validates the critical contribution of each module. Future work will extend validation to additional datasets and investigate the adaptation of the framework to simultaneous OC segmentation for comprehensive clinical risk assessment.
\bibliographystyle{IEEEtran}
\bibliography{strings}

@article{haleem2018novel,
  title={A novel adaptive deformable model for automated optic disc and cup segmentation to aid glaucoma diagnosis},
  author={Haleem, Muhammad Salman and Han, Liangxiu and Hemert, Jano van and Li, Baihua and Fleming, Alan and Pasquale, Louis R and Song, Brian J},
  journal={Journal of medical systems},
  volume={42},
  number={1},
  pages={20},
  year={2018},
  publisher={Springer}, 
  doi = {}
}

@article{adams1994seeded,
  title={Seeded region growing},
  author={Adams, Rolf and Bischof, Leanne},
  journal={IEEE Transactions on pattern analysis and machine intelligence},
  volume={16},
  number={6},
  pages={641--647},
  year={1994},
  publisher={IEEE}
}

@inproceedings{adhikari2015spatial,
  title={A spatial fuzzy C-means algorithm with application to MRI image segmentation},
  author={Adhikari, Sudip Kumar and Sing, Jamuna Kanta and Basu, Dipak Kumar and Nasipuri, Mita},
  booktitle={2015 Eighth International Conference on Advances in Pattern Recognition (ICAPR)},
  pages={1--6},
  year={2015},
  organization={IEEE}
}

@article{elazab2015segmentation,
  title={Segmentation of brain tissues from magnetic resonance images using adaptively regularized kernel-based fuzzy C-means clustering},
  author={Elazab, Ahmed and Wang, Changmiao and Jia, Fucang and Wu, Jianhuang and Li, Guanglin and Hu, Qingmao},
  journal={Computational and mathematical methods in medicine},
  volume={2015},
  number={1},
  pages={485495},
  year={2015},
  publisher={Wiley Online Library}
}

@article{thakur2019optic,
  title={Optic disc and optic cup segmentation from retinal images using hybrid approach},
  author={Thakur, Niharika and Juneja, Mamta},
  journal={Expert Systems with Applications},
  volume={127},
  pages={308--322},
  year={2019},
  publisher={Elsevier}
}

@article{chan2018role,
  title={The role of optical coherence tomography in the acute management of neuro-ophthalmic diseases},
  author={Chan, Noel CY and Chan, Carmen KM},
  journal={The Asia-Pacific Journal of Ophthalmology},
  volume={7},
  number={4},
  pages={265--270},
  year={2018},
  publisher={LWW}
}

@article{lalonde2002fast,
  title={Fast and robust optic disc detection using pyramidal decomposition and Hausdorff-based template matching},
  author={Lalonde, Marc and Beaulieu, Mario and Gagnon, Langis},
  journal={IEEE transactions on medical imaging},
  volume={20},
  number={11},
  pages={1193--1200},
  year={2002},
  publisher={IEEE}
}

@article{achanta2012slic,
  author       = {Radhakrishna Achanta and
                  Appu Shaji and
                  Kevin Smith and
                  Aur{\'{e}}lien Lucchi and
                  Pascal Fua and
                  Sabine S{\"{u}}sstrunk},
  title        = {{SLIC} Superpixels Compared to State-of-the-Art Superpixel Methods},
  journal      = {{IEEE} Trans. Pattern Anal. Mach. Intell.},
  volume       = {34},
  number       = {11},
  pages        = {2274--2282},
  year         = {2012},
  url          = {https://doi.org/10.1109/TPAMI.2012.120},
  doi          = {10.1109/TPAMI.2012.120},
  bibsource    = {dblp computer science bibliography, https://dblp.org}
}

@inproceedings{ronneberger2015u,
  title={U-net: Convolutional networks for biomedical image segmentation},
  author={Ronneberger, Olaf and Fischer, Philipp and Brox, Thomas},
  booktitle={International Conference on Medical image computing and computer-assisted intervention},
  pages={234--241},
  year={2015},
  organization={Springer}
}

@article{aquino2010detecting,
  title={Detecting the optic disc boundary in digital fundus images using morphological, edge detection, and feature extraction techniques},
  author={Aquino, Arturo and Geg{\'u}ndez-Arias, Manuel Emilio and Mar{\'\i}n, Diego},
  journal={IEEE transactions on medical imaging},
  volume={29},
  number={11},
  pages={1860--1869},
  year={2010},
  publisher={IEEE}
}

@article{Singhetal2020,
  author  = {Singh, Amitojdeep and Sengupta, Sourya and Lakshminarayanan, Vasudevan},
  title   = {Explainable Deep Learning Models in Medical Image Analysis},
  journal = {Journal of Imaging},
  year    = {2020},
  volume  = {6},
  number  = {6},
  pages   = {52},
  doi     = {10.3390/jimaging6060052}
}

@article{Chenetal2023,
  author  = {Chen, Yiyang and Ge, Pengqiang Susu and Wang, Guina and Weng, Guirong and Chen, Hongtian},
  title   = {An overview of intelligent image segmentation using active contour models},
  journal = {Intelligence \& Robotics},
  year    = {2023},
  volume  = {3},
  number  = {1},
  pages   = {23--55},
  doi     = {10.20517/ir.2023.02}
}

@article{Zhangetal2020,
  author    = {Zhang, Ling and Wang, Xiaosong and Yang, Dong and Sanford, Thomas and Harmon, Stephanie and Turkbey, Baris and Wood, Bradford J. and Roth, Holger R. and Myronenko, Andriy and Xu, Daguang},
  title     = {Generalizing Deep Learning for Medical Image Segmentation to Unseen Domains via Deep Stacked Transformation},
  journal = {IEEE Transactions on Medical Imaging},
  year      = {2020},
  note      = {ArXiv preprint arXiv:2002.07343}
}

@article{Kimetal2022,
  author  = {J. Kim and L. Tran and T. Peto and E. Y. Chew},
  title   = {Identifying Those at Risk of Glaucoma: A Deep Learning Approach for Optic Disc and Cup Segmentation and Their Boundary Analysis},
  journal = {Diagnostics},
  year    = {2022},
  volume  = {12},
  number  = {5},
  pages   = {1063},
  doi     = {10.3390/diagnostics12051063}
}

@article{cheng2013superpixel,
  title={Superpixel classification based optic disc and optic cup segmentation for glaucoma screening},
  author={Cheng, Jun and Liu, Jiang and Xu, Yanwu and Yin, Fengshou and Wong, Damon Wing Kee and Tan, Ngan-Meng and Tao, Dacheng and Cheng, Ching-Yu and Aung, Tin and Wong, Tien Yin},
  journal={IEEE transactions on medical imaging},
  volume={32},
  number={6},
  pages={1019--1032},
  year={2013},
  publisher={IEEE}
}

@inproceedings{long2015fully,
  title={Fully convolutional networks for semantic segmentation},
  author={Long, Jonathan and Shelhamer, Evan and Darrell, Trevor},
  booktitle={Proceedings of the IEEE conference on computer vision and pattern recognition},
  pages={3431--3440},
  year={2015}
}

@article{chen2017deeplab,
  title={Deeplab: Semantic image segmentation with deep convolutional nets, atrous convolution, and fully connected crfs},
  author={Chen, Liang-Chieh and Papandreou, George and Kokkinos, Iasonas and Murphy, Kevin and Yuille, Alan L},
  journal={IEEE transactions on pattern analysis and machine intelligence},
  volume={40},
  number={4},
  pages={834--848},
  year={2017},
  publisher={IEEE}
}

@incollection{Zuiderveld1994CLAHE,
  author    = {Zuiderveld, Karel},
  title     = {Contrast Limited Adaptive Histogram Equalization},
  booktitle = {Graphics Gems IV},
  editor    = {Heckbert, Paul S.},
  publisher = {Academic Press Professional, Inc.},
  year      = {1994},
  pages     = {474--485},
  isbn      = {0-12-336155-9},
  doi       = {10.1016/B978-0-12-336156-1.50061-6}
}

@article{Rother2004GrabCut,
  author    = {Rother, Carsten and Kolmogorov, Vladimir and Blake, Andrew},
  title     = {GrabCut: Interactive Foreground Extraction using Iterated Graph Cuts},
  journal   = {ACM Transactions on Graphics (TOG)},
  volume    = {23},
  number    = {3},
  pages     = {309--314},
  year      = {2004},
  publisher = {ACM New York, NY, USA},
  doi       = {10.1145/1015706.1015720}
}

@article{shahriari2016loop,
  author    = {Shahriari, Bobak and Swersky, Kevin and Wang, Ziyu and Adams, Ryan P. and de Freitas, Nando},
  title     = {Taking the Human Out of the Loop: A Review of Bayesian Optimization},
  journal   = {Proceedings of the IEEE},
  year      = {2016},
  volume    = {104},
  number    = {1},
  pages     = {148--175},
  doi       = {10.1109/JPROC.2015.2494218}
}

@inproceedings{Sivaswamy2014DrishtiGS,
  author    = {Sivaswamy, Jayanthi and Krishnadas, S. R. and Joshi, Gopal Datt and Jain, MD and Tabish, A. U. Syed},
  title     = {Drishti-GS: Retinal image dataset for optic nerve head (ONH) segmentation},
  booktitle = {2014 IEEE 11th International Symposium on Biomedical Imaging (ISBI)},
  year      = {2014},
  pages     = {53--56},
  doi       = {10.1109/ISBI.2014.6867807},
  organization = {IEEE}
}

\end{document}